\documentclass[10pt,twocolumn,letterpaper]{article}

\usepackage[pagenumbers]{cvpr} 

\usepackage{graphicx}
\usepackage{amsmath}
\usepackage{amssymb}
\usepackage{booktabs}
\usepackage{multirow}

\usepackage[pagebackref,breaklinks,colorlinks]{hyperref}

\usepackage[capitalize]{cleveref}
\crefname{section}{Sec.}{Secs.}
\Crefname{section}{Section}{Sections}
\Crefname{table}{Table}{Tables}
\crefname{table}{Tab.}{Tabs.}

\def\cvprPaperID{*****} 
\def\confName{CVPR}
\def\confYear{2022}

\begin{document}

\title{Subtask-dominated Transfer Learning for Long-tail Person Search}

\author{Chuang Liu~\hspace{10pt}
Hua Yang~\hspace{10pt}
Qin Zhou~\hspace{10pt}
Shibao Zheng \\
Shanghai Jiao Tong University\\
}
\maketitle

\begin{abstract}
Person search unifies person detection and person re-identification (Re-ID) to locate query persons from the panoramic gallery images. One major challenge comes from the imbalanced long-tail person identity distributions, which prevents the one-step person search model from learning discriminative person features for the final re-identification. However, it is under-explored how to solve the heavy imbalanced identity distributions for the one-step person search. Techniques designed for the long-tail classification task, for example, image-level re-sampling strategies, are hard to be effectively applied to the one-step person search which jointly solves person detection and Re-ID subtasks with a detection-based multi-task framework. To tackle this problem, we propose a Subtask-dominated Transfer Learning (STL) method. The STL method solves the long-tail problem in the pretraining stage of the dominated Re-ID subtask and improves the one-step person search by transfer learning of the pretrained model. We further design a Multi-level RoI Fusion Pooling layer to enhance the discrimination ability of person features for the one-step person search. Extensive experiments on CUHK-SYSU and PRW datasets demonstrate the superiority and effectiveness of the proposed method.
\end{abstract}

\section{Introduction}

\begin{figure}[!t]
 \centering
\begin{subfigure}{\linewidth}
  \includegraphics[width=1.0\linewidth]{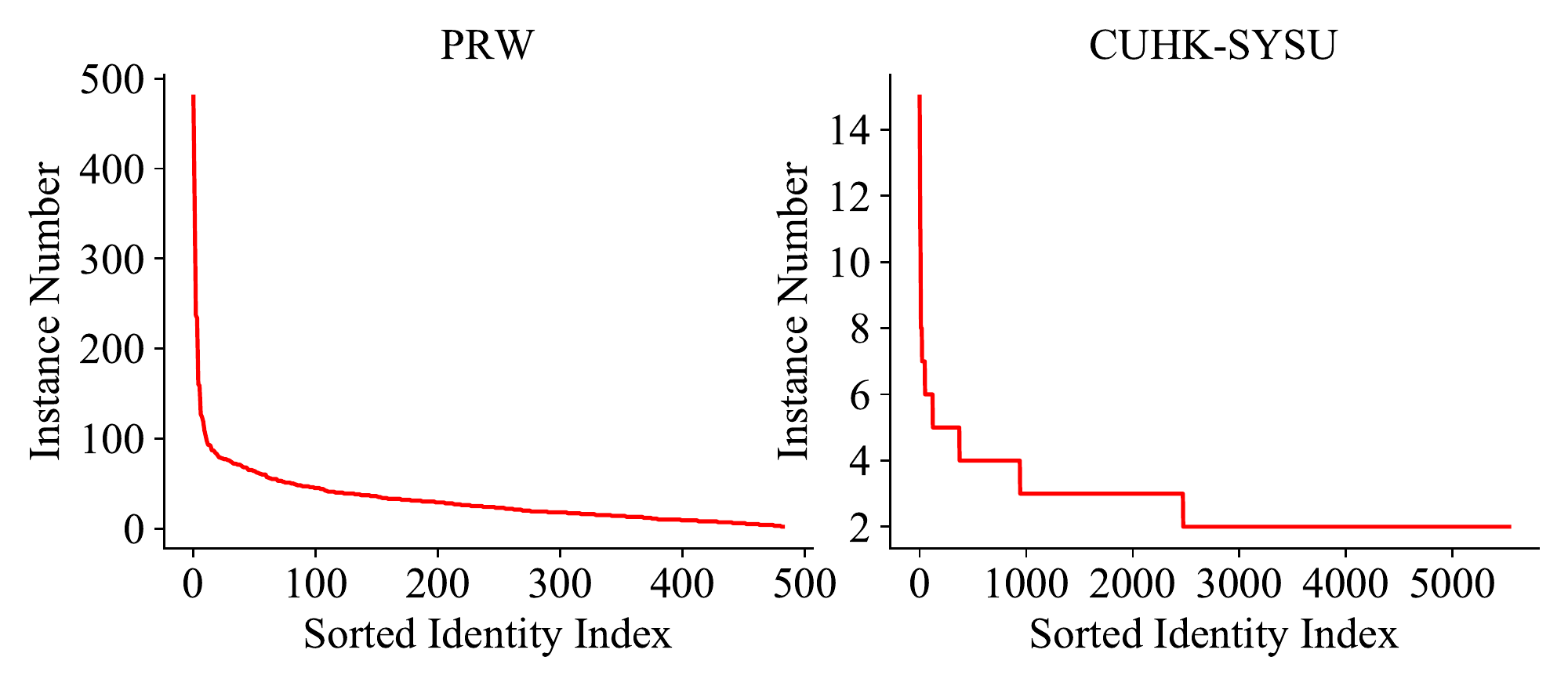}
  \caption{Long-tail distributions on person search datasets.}
    \label{fig:motivation-a}
  \end{subfigure}
 \begin{subfigure}{\linewidth}
  \includegraphics[width=1.0\linewidth]{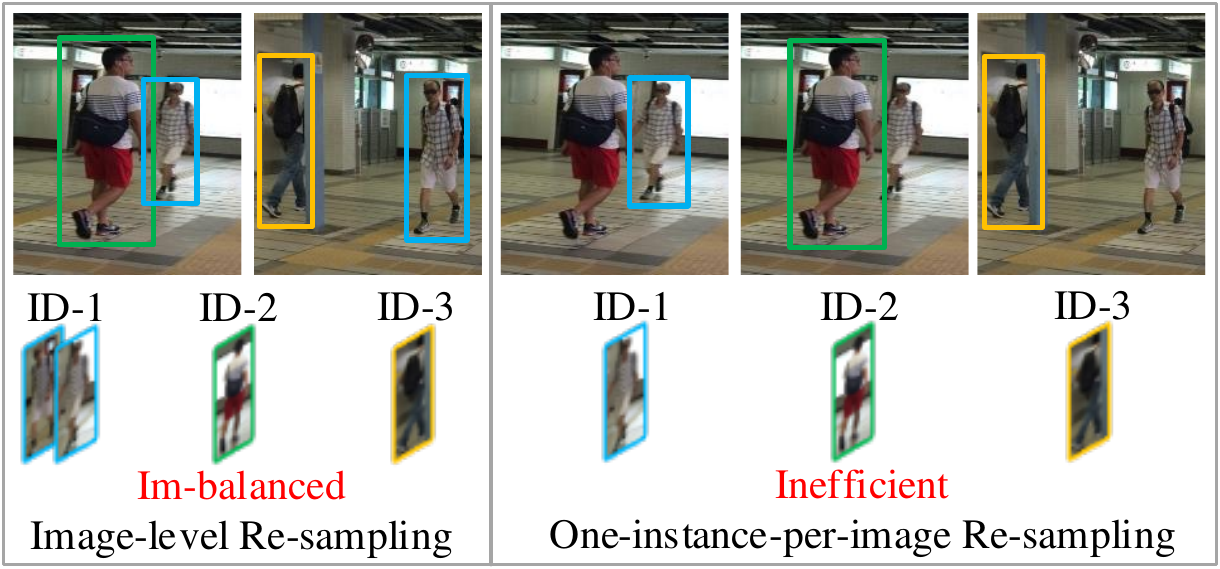}
  \caption{Re-sampling methods for one-step person search.}
    \label{fig:motivation-b}
  \end{subfigure}

\caption{Illustration of the long-tail challenge for one-step person search. (a) One-step person search suffers from the long-tail identity distribution problem. (b) The image re-sampling methods are hard to efficiently obtain balanced instance samples when applied to the one-step person search.}
\label{fig:motivation}
\end{figure}

Person search task aims to locate query persons from the panoramic scene images. Compared to the traditional person re-identification (Re-ID) task, person search task requires both person detection and person Re-ID, making it more applicable to real-world applications. Apart from the common challenges from person detection and person Re-ID, e.g. cluttered background, low resolution, occlusion, camera-view changes and person pose variations, person search also suffers from the heavy long-tail identity distributions, as shown in Figure~\ref{fig:motivation-a}. In this case, the tail identities (identities with few samples) are easily overwhelmed by the head identities (identities with plenty of samples) during the training phase, making the person search model hard to learn discriminative person features for re-identification. Although long-tail classification has emerged as a hot topic, it has been under-explored how to solve the long-tail distribution problem in the person search community. In this paper, we focus on addressing the long-tail identity distribution challenge for person search.

Most person search approaches can be divided into two categories, the two-step and the one-step. The two-step methods train a person detector and a person Re-ID model separately, and then cascade them to solve the person search problem, such as \cite{zheng2017person,chen2018person,wang2020tcts}. In contrast, the one-step methods solve the person search problem end-to-end with a detection-based multi-task framework, such as \cite{xiao2017joint,chen2020norm,dong2020bi}. Compared to the two-step framework, the one-step framework has fewer parameters and computations. Thus, we pay attention to solving the long-tail problem for the one-step person search in this paper.

A straightforward solution is to directly adapt long-tail classification techniques to the one-step person search, such as re-sampling training images~\cite{chawla2002smote,ando2017deep,pouyanfar2018dynamic} and re-weighting classification loss~\cite{zhou2010multi,lin2017focal,cao2019learning,huang2019deep}. The re-sampling methods usually conduct image-level re-sampling strategies to obtain balanced samples. However, as shown in Figure~\ref{fig:motivation-b}, for the one-step person search, it is inefficient to obtain balanced instance samples by performing an image-level re-sampling strategy on panoramic images containing multiple instances. Besides, the re-weighting methods are sensitive to hyper-parameters and hard to choose proper hyper-parameters to bring performance improvement for person search. Therefore, it is necessary to develop a simple yet effective method to solve the long-tail identity distribution problem for the one-step person search.

In this paper, we propose a Subtask-dominated Transfer Learning (STL) method to overcome the long-tail challenge for the end-to-end one-step person search. The STL method overcomes the long-tail identity distribution challenge in the pretraining stage of the dominated subtask and then transfers the pretrained model to the one-step person search model to reduce the negative influence of long-tail distributions. Specifically, the proposed STL method takes the Re-ID subtask as the dominated subtask and pretrains a Re-ID model first. Different from the one-step person search model, the pretrained Re-ID model takes as input the person instance images rather than the scene images, which makes it possible to effectively apply some long-tail classification techniques (e.g. the re-sampling methods) to the pretraining stage. Thus, the pretrained Re-ID model can generate discriminative person features for re-identification. Then, the backbone weights of the pretrained Re-ID model are utilized to initialize the one-step person search model to alleviate the influence of long-tail identity distributions. The local connectivity and parameter sharing of CNN guarantee the feasibility of this cross-task transfer learning. Finally, the one-step person search model can be directly trained to generate accurate detections and discriminative enough person features with detection losses for person detection and only cross-entropy loss for person Re-ID.

To further enhance the discrimination ability of the learned person features, we design a Multi-level RoI Fusion Pooling (MRFP) layer for the one-step person search. Specifically, for a predicted person Region of Interest (RoI), we propose to crop corresponding feature maps from multi-levels of the shared CNN parts and fuse the cropped multi-scale feature maps from different levels before feeding them to the following networks. Compared to the widely-used single-level RoI pooling operation, the proposed MRFP layer can keep more details which are helpful to distinguish different person identities for the identification networks, and consequently helps the identification networks learn more discriminative features for detected persons.

In summary, this paper makes the following contributions to the person search community:
\begin{itemize}
    \item We propose a simple yet effective method to solve the long-tail distribution problem for person search, namely the Subtask-dominated Transfer Learning (STL) method. The STL can greatly boost the one-step person search.
    \item We design the Multi-level RoI Fusion Pooling (MRFP) layer to further improve the discrimination ability of the learned person features by multi-level and multi-scale RoI feature pooling and fusion.
    \item We demonstrate the superiority of our proposed method through extensive experiments on two public person search datasets.
\end{itemize}

\section{Related Work}

\textbf{Person Search.} Recently, person search has received lots of attention from computer vision community researchers. Generally, there are two mainstream person search frameworks, the two-step and the one-step.

\cite{zheng2017person}~\cite{zheng2017person} propose the two-step framework and thoroughly evaluate combinations of different person detectors and person Re-ID models. Following the one-step framework,
\cite{chen2018person}~\cite{chen2018person} use the Faster R-CNN~\cite{ren2015faster} to detect persons and develop a two-stream CNN model to obtain representative features for persons by fusing the global features and local features. 
\cite{lan2018person}~\cite{lan2018person} propose the Cross-Level Semantic Alignment to solve the multi-scale challenge by combining cross-level feature maps. 
\cite{wang2020tcts}~\cite{wang2020tcts} propose a Task-Consist Two-Stage person search framework including an identity-guided query detector to generate query-like person detections and a Detection Results Adapted Re-ID model to make the Re-ID model adapted to the detections.

\cite{xiao2017joint}~\cite{xiao2017joint} propose the one-step framework based on the Faster R-CNN detection framework and design the OIM loss to tackle the ill-conditioned training problem. 
Following the one-step framework,
\cite{munjal2019query}~\cite{munjal2019query} propose a query-guided one-step person search model which can generate query-relevant proposals.
\cite{chen2020norm}~\cite{chen2020norm} propose the Norm-Aware Embedding method to solve the conflict between person detection and Re-ID by disentangling person embeddings into norms and angles to conduct detection and Re-ID, respectively.
\cite{dong2020bi}~\cite{dong2020bi} develop the Bi-directional Interaction Network to learn more discriminative person features by reducing the redundant information outside a bounding box.

\begin{figure*}[t]
\begin{center}
  \includegraphics[width=1.0\linewidth]{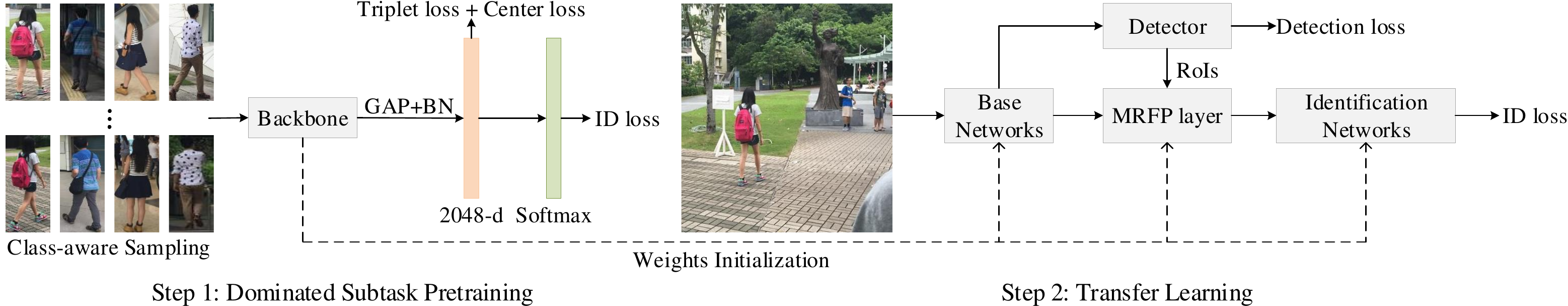}
\end{center}
\caption{Pipeline of the proposed method for the one-step person search framework.}
\label{fig:person_search_framework}
\end{figure*}

\textbf{Long-tail Classification.} Since real-world data is typically long-tailed with imbalanced class distributions, long-tail classification has raised more and more attention from researchers. Re-sampling and re-weighting are the mainstream solutions to the long-tail classification problem. The re-sampling methods either over-sample the tail classes~\cite{chawla2002smote,ando2017deep,pouyanfar2018dynamic}, or under-sample the head classes~\cite{drumnond2003class,buda2018systematic}, or conduct class-aware sampling~\cite{shen2016relay}. For the re-weighting methods, an intuitive strategy is to re-weight the sample losses at class level according to the class frequency~\cite{mikolov2013distributed,wang2017learning}. Besides the class-level re-weighting methods, some works focus on re-weighting losses at sample level based on the difficulty level of samples~\cite{lin2017focal,li2019gradient}. In addition, some researchers also try to solve the long-tail distribution problem by metric learning~\cite{huang2016learning,zhang2017range} or self-supervised pretraining transfer learning~\cite{yang2020rethinking}.

\section{Method}

The overall pipeline of the proposed STL method is shown in Figure~\ref{fig:person_search_framework}. In the STL method, the backbone is first pretrained on the dominated Re-ID subtask and then transferred to the one-step person search model. For the one-step person search model, we adopt the framework proposed by~\cite{liu2021making}~\cite{liu2021making} with the  ResNet50-IBN-a~\cite{pan2018two} as the backbone model. The backbone is divided into two parts, namely the base networks and identification networks. Given an input scene image, the base networks first extract its convolutional (conv) feature maps. Then, the detector predicts the possible person RoIs based on the conv feature maps from the base networks. Next, for each predicted person RoI, the proposed MRFP layer conducts multi-level multi-scale RoI feature maps cropping and fusion operations to generate the RoI input feature maps for the following identification networks. Finally, the identification networks extract the final feature for each predicted person RoI. In the following sections, the proposed STL and the MRFP layer are introduced in detail.

\subsection{Subtask-dominated Transfer Learning}

\begin{figure}[t]
\begin{center}
  \includegraphics[width=1.0\linewidth]{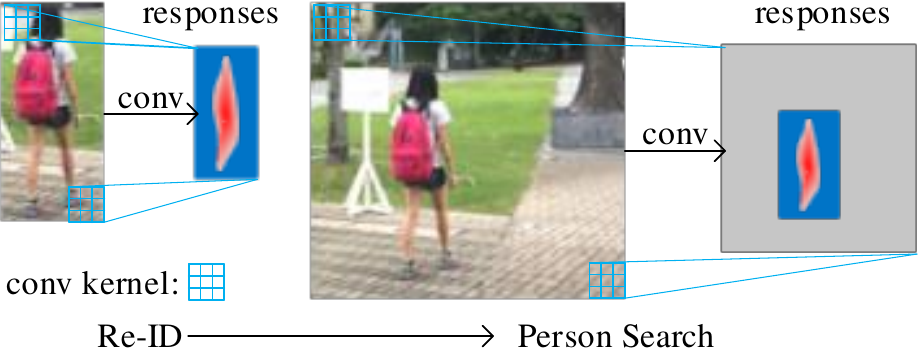}
\end{center}
\caption{Illustration of applying conv kernels trained on Re-ID subtask to the one-step person search.}
\label{fig:conv}
\end{figure}

In the one-step person search, the person detection subtask focuses on the differences between foreground persons and background, and the person Re-ID subtask focuses on the differences among different person identities. Thus, based on the Re-ID-dominated feature maps, a CNN model can still distinguish the foreground persons from background by learning the commonness of foreground persons. Therefore, it does not bring negative influences on the person detection subtask to take the Re-ID subtask as the dominated subtask of the one-step person search.
Besides, as shown in Figure~\ref{fig:conv}, due to the local connectivity and parameter sharing of conv, conv kernels trained on person patches can still generate similar responses in person RoIs when applied to scene images. Based on these analyses, we propose the STL method to solve the long-tail problem for the one-step person search.

Specifically, the proposed STL method includes two steps, dominated subtask pretraining and transfer learning. Details are as follows:

\textbf{Step 1: Dominated Subtask Pretraining.} The long-tail identity distribution problem prevents the Re-ID subtask from learning discriminative person features. Thus, the Re-ID subtask is chosen as the dominated subtask for the one-step person search. In the one-step person search model shown in Figure~\ref{fig:person_search_framework}, networks related to Re-ID subtask are the base networks and identification networks. These two networks form a complete ResNet50-IBN-a backbone model. Thus, in this step, a backbone model is pretrained. The pipeline proposed by~\cite{luo2019bag}~\cite{luo2019bag}, shown in Figure~\ref{fig:person_search_framework}, is adopted to pretrain the backbone model for the dominated Re-ID subtask. 

To pretrain the backbone model, a Re-ID style training dataset is constructed by cropping all the instances from the original scene images in the training set. All cropped instances are resized to $256\times 128$ ones. All training images on the Re-ID style training set are single-instance-level and the pipeline to pretrain the backbone model is also single-task, which makes the re-sampling methods and metric learning methods requiring special sampling strategies easily adapted to the pretraining of the backbone model. Thus, we adopt the re-sampling strategy and metric learning methods for the pretraining of the backbone model to relieve the influence of the long-tail identity distribution.

For the former, the class-aware sampling strategy~\cite{shen2016relay} is performed. Specifically, $P$ identities are first randomly selected, and then $K$ instances are randomly chosen for each selected identity to construct a mini-batch with size $P\times K$. For an identity, if the number of instances is smaller than $K$, random repeated sampling is conducted to guarantee $K$ instances. For the latter, the batch hard triplet loss~\cite{hermans2017defense} and center loss~\cite{wen2016discriminative} are employed. The batch hard triplet loss also requires the class-aware sampling. As shown in Figure~\ref{fig:person_search_framework}, the batch hard triplet loss and center loss are computed based on the 2048-d embeddings. Besides, a softmax layer is finally applied to compute the cross-entropy loss as the ID loss. The final loss $L_{pre}$ used to pretrain the backbone model is computed as follows:
\begin{equation}
    L_{pre} = L_{id}+L_{tri}+\lambda \cdot L_{center}
    \label{eq.1}
\end{equation}
where $L_{id}$, $L_{tri}$, $L_{center}$ and $\lambda$ are the ID loss, batch hard triplet loss, center loss and weight factor, respectively. 

Intuitively, the more powerful the pretrained backbone model is, the better the transfer learning result is. Therefore, some useful training tricks collected by \cite{luo2019bag}~\cite{luo2019bag} are adopted to pretrain a powerful backbone model, including random erasing augmentation and learning rate warming up.

\textbf{Step 2: Transfer Learning.} To transfer the knowledge of the pretrained Re-ID model to the one-step person search model, we adopt the pretraining and fine-tuning transfer learning method. Specifically, the weights of the pretrained Re-ID backbone are used to initialize the backbone of the one-step person search model. Then, we fine-tune the one-step person search model to make it adapted to the person search task. 

Following the Faster R-CNN~\cite{ren2015faster}, we employ the RPN training losses ($L_{cls}^{\text{rpn}}$ and $L_{reg}^{rpn}$) and RoI Head training losses ($L_{cls}$ and $L_{reg}$) to train the person detector. The total detection loss $L_{det}$ is defined as follows:
\begin{equation}
    L_{det}=L_{cls}^{rpn}+L_{reg}^{rpn}+L_{cls}+L_{reg}.
    \label{eq.2}
\end{equation}
The cross-entropy loss is computed as the ID loss $L_{id}$ to train the identification networks. Overall, the total loss to train the one-step person search model is defined as follows:
\begin{equation}
    L=L_{id}+L_{det}.
    \label{eq.3}
\end{equation}

\subsection{Multi-level RoI Fusion Pooling}

In the previous one-step person search model, the RoI pooling operation is conducted to crop RoI feature maps only from the output of the base networks. However, the base networks are shared by both the person detection and Re-ID subtasks. Influenced by the person detection subtask, some important details which can help to distinguish different persons are likely to be missing in the high-level output of the base networks, which harms the discrimination ability of person features learned by the identification networks. To address this issue, we propose the Multi-level RoI Fusion Pooling (MRFP) layer to replace the widely-used single-level RoI pooling. Since the low-level feature maps contain more details, the proposed MRFP layer crops the RoI feature maps from multi-levels of the base networks and fuses the cropped multi-level multi-scale feature maps to keep more person details in the pooled feature maps.

\begin{figure}[t]
\begin{center}
  \includegraphics[width=1.0\linewidth]{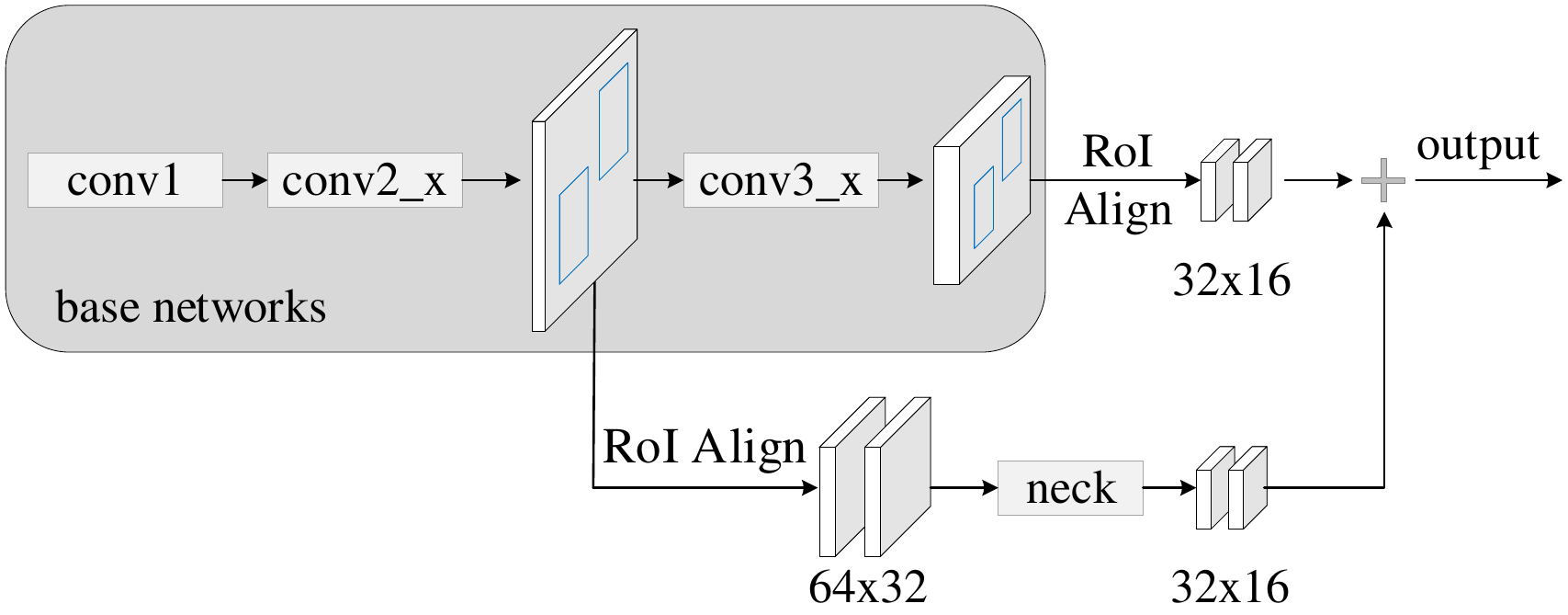}
\end{center}
\caption{Illustration of the proposed MRFP layer.}
\label{fig:MRFP}
\end{figure}

As shown in Figure~\ref{fig:MRFP}, based on the output feature maps of the high-level conv3\_x, the RoI Align~\cite{he2017mask} pooling is conducted to pool the RoI feature maps into $32\times 16$ ones. Besides, in the proposed MRFP layer, the RoI Align is also performed to the output feature maps of the low-level conv2\_x. Since the output resolution of conv2\_x is twice as large as that of conv3\_x, the pooling size for conv2\_x is $64\times 32$. To perform the fusion of RoI feature maps from two levels, a neck layer is applied to transform the low-level RoI feature maps into the same size RoI feature maps as the high-level ones. In this work, we use the conv3\_x of the backbone as the neck layer to keep semantic consistency between the transformed RoI feature maps and the high-level ones. Please note that the neck layer does not share parameters with the conv3\_x in the base networks. Then, the pixel-wise sum is utilized to fuse RoI feature maps from two levels. The fused RoI feature maps are to be fed into the identification networks to extract person features.

\subsection{Architecture Details}
In the person search model, the base networks contain layers from conv1 to conv3\_x of the backbone, and the identification networks contain layers from conv4\_x and conv5\_x of the backbone. Additionally, in the identification networks, we modify the convolution stride of conv5\_x to 1 and add a Global Average Pooling (GAP) layer followed by a Batch Normalization (BN) layer to obtain the embedding. 

For the detector, we adopt the Faster R-CNN framework~\cite{ren2015faster} which includes a standard RPN and a RoI Head. Specifically, the RPN first uses a conv layer (kernel size: $3\times3$, stride: 1) to process the input feature maps, and adopts another two conv layers (kernel size: $1\times 1$, stride: 1) to classify the RoI proposals and predict their coordinates. Based on the RoI proposals from the RPN module, the RoI Align operation is used to crop RoI feature maps into $7\times 7$ ones. The cropped feature maps are flattened to vectors before being fed into the RoI Head. The RoI Head is composed of two cascaded 1024-D fully-connected layers followed by a box-classification layer and a box-regression layer.

\section{Experiments}

In this section, we run experiments on two public person search datasets, the PRW~\cite{zheng2017person} and CUHK-SYSU~\cite{xiao2017joint}, and compare the proposed method with some state-of-the-art methods. Afterward, ablation study results are reported for each component in the proposed method.

\subsection{Datasets}

\textbf{PRW} dataset~\cite{zheng2017person} is collected in Tsinghua university by six cameras. A total of 11,816 video frames containing 43,110 person bounding boxes are provided. The training set includes 5,704 frames where 15,575 person bounding boxes are labeled with 482 identities and the rest bounding boxes are unlabeled. For the testing set, 2,057 labeled person bounding boxes are marked as the query set, and 6,112 frames are taken as the gallery set. The search scope is the whole gallery set.

\textbf{CUHK-SYSU} dataset~\cite{xiao2017joint} collects video frames from the street snap and movies. A total of 18,184 frames with 96,143 person bounding boxes are provided. The training set includes 11,206 frames containing 15,080 person bounding boxes labeled with 5,532 identities and a lot of bounding boxes without identity labels. The testing set is composed of 2,900 labeled query persons and 6,987 gallery frames. Different from the PRW dataset, for each query person, the CUHK-SYSU dataset provides several gallery subsets with various gallery sizes.

\subsection{Evaluation Protocols} 

The Cumulative Matching Characteristic (CMC top-1) and mean Average Precision (mAP) are adopted to evaluate the performance of person search. These two metrics are also the widely-used evaluation protocols in the person Re-ID area. However, different from Re-ID, the AP of each query person is scaled by its recall rate, and the mAP is calculated as the average of all APs across all query persons.

\subsection{Implementation Details}

For the backbone model, we pretrain it for total 120 epochs on the constructed Re-ID style training set. For class-aware sampling, on the PRW dataset, we first randomly select $P=16$ identities and choose $K=4$ instances for each identity to obtain a balanced batch with size 64. Since the total number of identities on the CUHK-SYSU is much larger than that on the PRW dataset, we set $P=64$ and $K=4$ to obtain a balanced batch with size 256 for better convergence during the training phase.
The Adam optimizer is employed to pretrain the backbone. The initial learning rate is $3.5\times  10^{-4}$ and decayed by a factor of 10 in 40-th and 70-th epochs, respectively. The learning rate is warmed up linearly from $3.5\times  10^{-5}$ to $3.5\times  10^{-4}$ in the first 10 epochs. The weight decay factor for the Adam optimizer is $5\times 10^{-4}$. The weight factor $\lambda$ in Eq.~\ref{eq.1} is set to $5 \times 10^{-4}$.
The random horizontal flip and random erasing~\cite{luo2019bag} are employed to augment the training data.

For the person search model, we train it end-to-end for total 20 epochs using the SGD optimizer with batch size 16. The learning rate is set to 0.005 initially and decayed to 0.0005 after 12 epochs. The weight decay factor for the SGD optimizer is $1\times 10^{-4}$.
Only the random horizontal flip data augmentation is used during the training of the person search model. 
For the RPN in the detector, the anchor sizes are set to 4, 8, 16, and 32 for each location on feature maps, and the anchor aspect ratios are set to 1, 2, and 3.
The height and width of an input scene image are scaled by the same factor to make the shorter side not less than 640 pixels or the longer side not more than 960 pixels.
During the reference phase, the predicted person bounding boxes with foreground scores lower than 0.5 are removed, and only bounding boxes whose Intersection over Union (IoU) with ground truth bounding boxes larger than 0.5 are regarded as true detection results.
All experiments are conducted on Pytorch. Source codes will be released upon acceptance.

\begin{table}[t]
\centering
\resizebox{0.45\textwidth}{!}{%
\begin{tabular}{@{}llccccc@{}}
\toprule
\multirow{2}{*}{Method} &   & \multicolumn{2}{c}{PRW}   & \multicolumn{2}{c}{CUHK-SYSU} \\  
\cmidrule(lr){3-4} \cmidrule(l){5-6} 
              &          & \multicolumn{1}{c}{mAP (\%)} & \multicolumn{1}{c}{top-1 (\%)} & mAP (\%)     & top-1 (\%)         \\ \midrule
\multicolumn{1}{c|}{\multirow{6}{*}{\rotatebox{90}{two-step}}}  & DPM+IDE~\cite{zheng2017person}  & 20.5      & 48.3    & -         & -        \\
\multicolumn{1}{c|}{}                                           & MGTS~\cite{chen2018person}      & 32.6      & 72.1    & 83.0      & 83.7     \\
\multicolumn{1}{c|}{}                                           & CLSA~\cite{lan2018person}       & 38.7      & 65.0    & 87.2      & 88.5     \\
\multicolumn{1}{c|}{}                                           & RDLR~\cite{han2019re}           & 42.9      & 70.2    & 93.0      & 94.2     \\
\multicolumn{1}{c|}{}                                           & IGPN~\cite{dong2020instance}    & 47.2      & 87.0    & 90.3      & 91.4     \\
\multicolumn{1}{c|}{}                                           & TCTS~\cite{wang2020tcts}        & \bf46.8   & \bf87.5 & \bf93.9   & \bf95.1  \\ 
\midrule 
\multicolumn{1}{c|}{\multirow{11}{*}{\rotatebox{90}{one-step}}} & OIM~\cite{xiao2017joint}        & -         & -       & 75.7      & 78.7     \\
\multicolumn{1}{c|}{}                                           & NPSM\cite{liu2017neural}        & 24.2      & 53.1    & 77.9      & 81.2     \\
\multicolumn{1}{c|}{}                                           & RCAA~\cite{chang2018rcaa}       & -         & -       & 79.3      & 81.3    \\
\multicolumn{1}{c|}{}                                           & IAN~\cite{xiao2019ian}          & 23.0      & 61.9    & 76.3      & 80.1     \\
\multicolumn{1}{c|}{}                                           & LCGPS~\cite{yan2019learning}    & 33.4      & 73.6    & 84.1      & 86.5     \\
\multicolumn{1}{c|}{}                                           & QEEPS~\cite{munjal2019query}    & 37.1      & 76.7    & 88.9      & 89.1     \\
\multicolumn{1}{c|}{}                                           & NAE+~\cite{chen2020norm}        & 44.0      & 81.1    & 92.1      & 92.9     \\
\multicolumn{1}{c|}{}                                           & APNet~\cite{zhong2020robust}    & 41.9      & 81.4    & 88.9      & 89.3     \\
\multicolumn{1}{c|}{}                                           & BINet~\cite{dong2020bi}         & 45.3      & 81.7    & 90.0      & 90.7     \\ 
\multicolumn{1}{c|}{}                                           & PSFL~\cite{kim2021prototype}    & 44.2      & 85.2    & 92.3      & \bf94.7  \\ 
\multicolumn{1}{c|}{}                                           & DKD~\cite{zhang2021diverse}     & 50.5      & 87.1    & 93.1      & 94.2     \\
\multicolumn{1}{c|}{}                                           & AlignPS~\cite{yan2021anchor}    & 45.9      & 81.9    & 93.1      & 93.4     \\ 
\multicolumn{1}{c|}{}                                           & Ours                            & \bf61.6   & \bf90.5 & \bf93.1   & 94.2     \\ 
\bottomrule
\end{tabular}%
}
\caption{Comparison with the state-of-the-art methods on the PRW and CUHK-SYSU datasets.}
\label{tab:comparison with SOTA}
\end{table}

\subsection{Comparison with State-of-the-art Methods}

In this section, we run experiments on the PRW and CUHK-SYSU datasets and compare the proposed person search method with some state-of-the-art methods.

\textbf{Comparison on PRW.} As shown in Table~\ref{tab:comparison with SOTA}, our proposed method achieves 61.6\% mAP and 90.5\% top-1 accuracy, outperforming all the compared methods by large margins. In the compared one-step methods, the IGPN method~\cite{dong2020instance} obtains the highest performance with 47.2\% mAP and 87.0 \% top-1. Compared to the IGPN method, our method surpasses it by 14.4\% mAP and 3.5\% top-1 accuracy. In the compared two-step methods, the strongest TCTS~\cite{wang2020tcts} method achieves 46.8\% mAP and 87.5\% top-1 accuracy. Compared with the TCTS method, our method obtains 14.8\% mAP and 3.0\% top-1 improvement. These comparison results demonstrate that our proposed method is superior to the compared state-of-the-art methods on the PRW dataset.

\textbf{Comparison on CUHK-SYSU.} The results on the CUHK-SYSU dataset are reported in Table~\ref{tab:comparison with SOTA}. Our proposed method achieves 93.1\% mAP and 94.2\% top-1 accuracy, surpassing most compared methods. It is observed that methods TCTS~\cite{wang2020tcts} and PSFL~\cite{kim2021prototype} obtain higher performance than ours. The TCTS method is two-step, requiring two independent models (a person detector and a person Re-ID model) to tackle the person search problem. In contrast, our method is one-step and solves the person search problem in a multi-task framework with fewer computations. The PSFL method designs a Prototype Guided Attention Module for saliency feature learning in the one-step person search framework. Even if our method does not contain an attention module for saliency feature learning, it still obtains comparable performance with the PSFL method. 

\begin{figure}[t]
\begin{center}
  \includegraphics[width=1.0\linewidth]{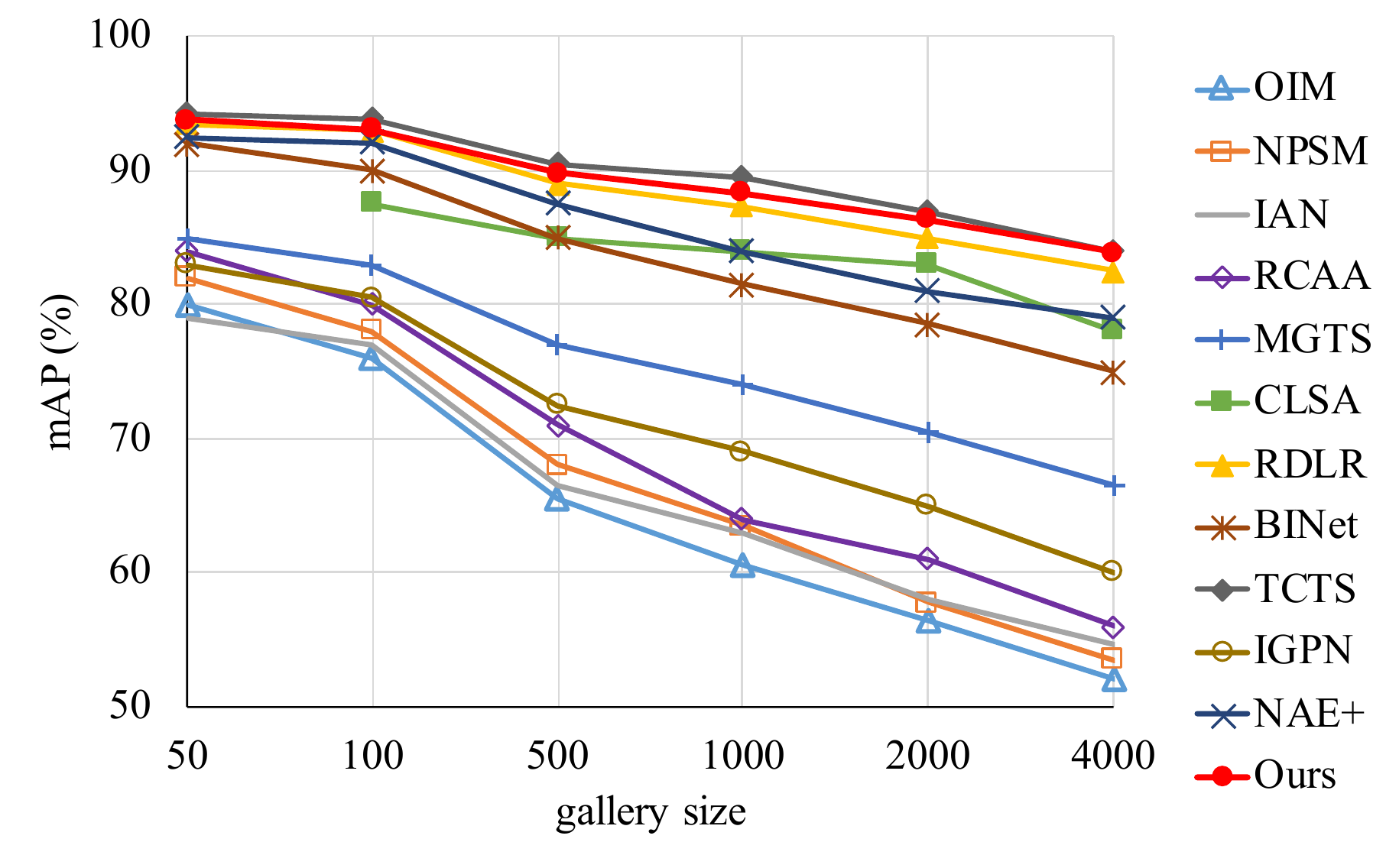}
\end{center}
\caption{Comparison on the CUHK-SYSU dataset with different gallery sizes.}
\label{fig:gallery_sizes}
\end{figure}

Besides, experiments are also conducted to explore the influences of different gallery sizes. The gallery size ranges from 50 to 4,000. As shown in Figure~\ref{fig:gallery_sizes}, the performance of all methods is degraded with the gallery size increasing, which indicates that it is still challenging to search for the target persons from the large search scope in real-world applications. However, when the gallery size increases, the proposed method outperforms all the compared state-of-the-art methods except a two-step method, the TCTS. Besides, it is observed that the proposed method obtains the same performance as the TCTS method for the largest gallery size 4,000. These experimental results demonstrate that the proposed method is more robust against gallery size variations and has advantages for large-scale person search.

\begin{table}[t]
\centering
\resizebox{0.35\textwidth}{!}{%
\begin{tabular}{@{}lcc@{}}
\toprule
{Method}                 & {mAP (\%)} & {top-1 (\%)}      \\ \midrule
Baseline (Random init)   &  29.3     &  77.0      \\
Baseline (ImageNet)      &  50.6     &  86.6      \\  
Baseline (STL)           &  \bf57.1  &  \bf88.3    \\ \midrule
Baseline (ImageNet)+MRFP &  53.0     &  87.7      \\
Baseline (STL)+MRFP      &  \bf61.6  &  \bf90.5   \\ \midrule
\end{tabular}%
}
\caption{Effectiveness of each proposed component. The ``Baseline (Random init)" means the backbone model is randomly initialized.}
\label{tab:effectiveness}
\end{table}

\subsection{Ablation Study}
In this section, experiments are conducted on the PRW dataset to validate the effectiveness of each proposed component. To study the influence of each component, we replace the proposed MRFP layer in the person search framework shown in Figure~\ref{fig:person_search_framework} with a RoI Align pooling layer to construct a baseline model. The baseline model is initialized with the ImageNet pretrained weights and denoted as ``Baseline (ImageNet)".

\textbf{Effectiveness of STL.}
In the previous person search works, the ImageNet pretraining is the most popular transfer learning way. To validate the effectiveness of the proposed STL, we compare it with the baseline model adopting ImageNet pretraining. As shown in Table~\ref{tab:effectiveness}, the ``Baseline (STL)" model achieves much higher performance than the ``Baseline (ImageNet)" model. Specifically, the ``Baseline (STL)" model outperforms the ``Baseline (ImageNet)" model by 6.5\% in mAP and 1.7\% in top-1. Besides, the ``Baseline (STL)+MRFP" surpasses the ``Baseline (ImageNet)+MRFP" by large margins.

Moreover, we also apply the proposed STL to some other one-step person search models to further validate its effectiveness, including the OIM model~\cite{xiao2017joint}, BINet~\cite{dong2020bi} and DKD~\cite{zhang2021diverse}. As shown in Table~\ref{tab:stl on other models}, when the STL method is applied, the person search performance is further improved compared to the ImageNet initialization method.
These experimental results further demonstrate the effectiveness of the proposed STL method.

Although ImageNet pretraining transfer learning is an effective method to improve the performance of the target task compared to the random initialization of target model weights, it is unrelated or weakly related to the target task, namely the person search task. Different from it, the proposed STL method is specially designed to address the long-tail problem for person search and is strongly related to the target task. Compared to the ImageNet pretraining, the proposed STL can provide a much better initial solution for the one-step person search. Consequently, the proposed STL method can better improve person search performance.

\begin{table}[t]
\centering
\resizebox{0.3\textwidth}{!}{%
\begin{tabular}{@{}lcc@{}}
\toprule
Method                   & {mAP (\%)}     & {top-1 (\%)}      \\ \midrule
OIM* (ImageNet)          &  42.8          &  82.7             \\
OIM* (STL)               &  \bf46.5       & \bf85.2           \\ \midrule
BINet* (ImageNet)        &  39.2          &  81.1             \\
BINet* (STL)             &  \bf43.4       &  \bf82.8          \\ \midrule
DKD* (ImageNet)          &  52.5          &  85.6             \\
DKD* (STL)               &  \bf54.0       &  \bf90.9          \\ \bottomrule
\end{tabular}%
}
\caption{Results of applying the STL to some other one-step person search models.}
\label{tab:stl on other models}
\end{table}

\textbf{Effectiveness of MRFP.}
As shown in Table~\ref{tab:effectiveness}, person search performance is further improved when the MRFP layer is applied. Specifically,  the ``Baseline (STL)+MRFP" outperforms the ``Baseline (STL)" by 4.5\% in mAP and 2.2\% in top-1. In addition, when the MRFP layer is applied to the ``Baseline (ImageNet)", person search performance is also improved, further validating the effectiveness of the MRFP layer.

\textbf{Applying re-weighting methods to person search.} Since the re-weighting long-tail methods are usually sensitive to hyper-parameters, it is hard to choose proper hyper-parameters to improve person search performance. To validate this, we compare the class-balanced cross-entropy loss~\cite{cui2019class} and focal loss~\cite{lin2017focal} with suggested hyper-parameters to the cross-entropy loss. Experimental results reported in Table~\ref{tab:re-weighting loss} shows that both the class-balanced cross-entropy loss and focal loss fail to bring performance improvement compared to the cross-entropy loss.

\textbf{Impact on detection.}
The STL method and MRFP layer are proposed to improve person search performance. Experiments are conducted to explore their impact on person detection subtask. Experimental results are reported in Table~\ref{tab:impact_on_detection}. It is observed that the ``Baseline (STL)" model obtains almost the same detection performance as the ``Baseline (ImageNet)" model. Moreover, the person search performance of ``Baseline (STL)+MRFP" is also nearly the same as ``Baseline (ImageNet)+MRFP". These comparisons show that our STL method almost has no negative influence on person detection performance and demonstrate that it is feasible to take the person Re-ID as the dominated subtask for the multi-task one-step person search framework. 

Besides, compared with ``Baseline (ImageNet)", ``Baseline (ImageNet)+MRFP" achieves comparable person detection performance. Additionally, ``Baseline (STL)+MRFP" also achieves almost the same person detection performance compared with ``Baseline (STL)". These experimental results validate that the proposed MRFP layer also has little impact on person detection performance.
\begin{table}[t]
\centering
\resizebox{0.4\textwidth}{!}{%
\begin{tabular}{@{}lcc@{}}
\toprule
{Method}                           &  mAP (\%)  & top-1 (\%)    \\ \midrule
Cross-entropy loss                 &  50.6     &  86.6          \\  
Class-balanced cross-entropy loss  &  48.6     &  85.8          \\
Focal loss                         &  47.7     &  85.8          \\ \midrule
\end{tabular}%
}
\caption{Results of directly applying the re-weighting loss to the one-step person search.}
\label{tab:re-weighting loss}
\end{table}
\begin{table}[t]
\centering
\resizebox{0.4\textwidth}{!}{%
\begin{tabular}{@{}lcc@{}}
\toprule
{Method}                      &  AP (\%)  & Recall (\%)    \\ \midrule
Baseline (ImageNet)           &  95.7     &  93.6          \\  
Baseline (STL)                &  95.6     &  93.5          \\
Baseline (ImageNet)+MRFP      &  95.2     &  93.1          \\  
Baseline (STL)+MRFP           &  95.3     &  93.3          \\ \midrule
\end{tabular}%
}
\caption{Person detection performance evaluation.}
\label{tab:impact_on_detection}
\end{table}

\textbf{Better pretrained models for better person search.}
To address the long-tail problem, the re-sampling method (Class-aware Sampling) and some metric learning methods (Batch Hard Triplet loss and Center loss) are adopted to pretrain the Re-ID model. An intuitive question is whether a better pretrained Re-ID model can help to obtain a better one-step person search model. To answer this question, we pretrain multiple Re-ID models under different training settings and evaluate the influences of these pretrained Re-ID models on the performance of the one-step person search model. To evaluate the performance of the pretrained Re-ID models, the Re-ID style testing datasets are constructed based on the annotated person bounding boxes and identities. Please kindly note that the class-aware sampling is only applied when the batch hard triplet loss is employed. 

Experimental results of the pretrained Re-ID model and one-step person search model are shown in Table~\ref{tab:reid_ps_performance}. It is observed that the performance of the pretrained model is further improved when the batch hard triplet loss and center loss are applied. Moreover, the performance of the one-step person search model is also improved with the enhancement of the pretrained Re-ID model. This indicates that the proposed STL method can further improve the one-step person search model by pretraining a stronger Re-ID subtask model.

\begin{table}[tp]
\centering
\resizebox{0.45\textwidth}{!}{%
\begin{tabular}{@{}llcc@{}}
\toprule
{Training Setting} &      & {mAP (\%)} & {top-1 (\%)}    \\ \midrule
\multicolumn{1}{l|}{\multirow{3}{*}{Re-ID}}&Softmax    &  60.7     &  78.2    \\  
\multicolumn{1}{l|}{}  & Softmax+Triplet               &  63.5     &  79.7    \\
\multicolumn{1}{l|}{}  & Softmax+Triplet+Center        &  \bf65.1  &  \bf80.4 \\ \midrule  
\multicolumn{1}{l|}{\multirow{3}{*}{Person Search}}&Softmax   &  60.6     &  89.6   \\  
\multicolumn{1}{l|}{}  & Softmax+Triplet               &  61.3     &  90.0   \\
\multicolumn{1}{l|}{}  & Softmax+Triplet+Center        &  \bf61.6     &  \bf90.5  \\ 
\midrule
\end{tabular}%
}
\caption{Performance evaluation on the Re-ID style and person search style testing sets under various training settings.}
\label{tab:reid_ps_performance}
\end{table}

\begin{figure}[t]
\begin{center}
  \includegraphics[width=1\linewidth]{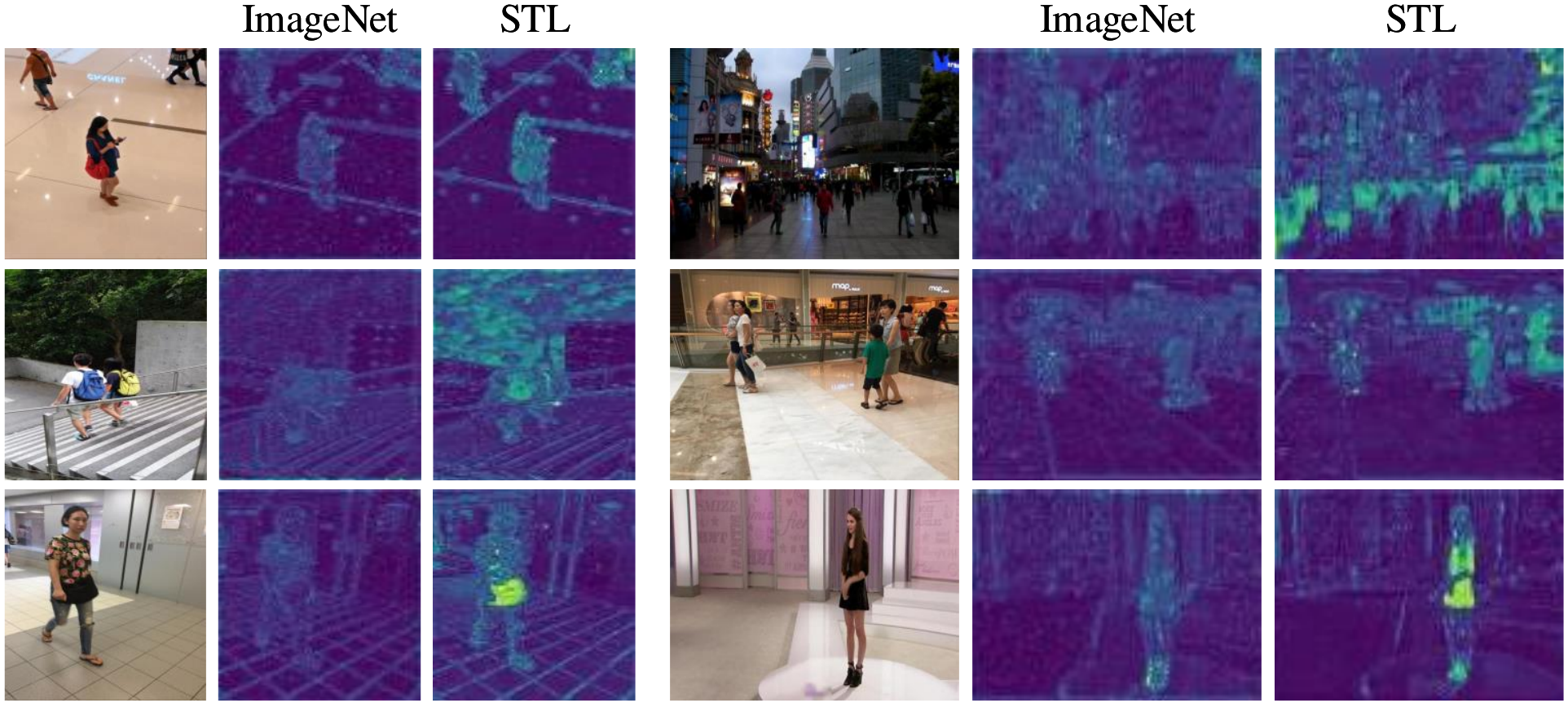}
\end{center}
\caption{Initial feature maps visualization of the ImageNet pretraining method and the proposed STL method.}
\label{fig:conv_visualization}
\end{figure}

\textbf{Visualization of Initial Feature Maps.}
We compare the ImageNet pretraining method and the proposed STL method by visualizing the initial feature maps of the base networks. As shown in Figure~\ref{fig:conv_visualization}, person areas of the output feature maps in the proposed STL method are brighter than those in the ImageNet pretraining method, which indicates that the proposed STL method can make the one-step person search model focus more on the person RoIs in the scene images compared to the ImageNet pretraining method before the training of the one-step person search model. Consequently, the proposed STL method can help the one-step person search model learn more discriminative person features after the training.  

\textbf{Running speed.} We compare the running speed of our method with some state-of-the-art methods. As shown in Table~\ref{tab:running_speed}, for the input image size $960\times 640$, the proposed ``Baseline (STL)+MRFP" method runs at a speed of 40.6 frames per second (fps), which is much faster than all the compared methods.
For far comparison, besides reporting the running speed on the $960\times 640$ input image size, we also report the running speed on the $1500\times 900$ input images size. For the $1500\times 900$ input image size, the proposed method ``Baseline (STL)" runs at a speed of 30.5 fps, faster than any compared method. When the proposed MRFP layer is applied, the running speed is slightly reduced to 28.7 fps, which is acceptable as the proposed MRFP layer contains a CNN neck layer. Totally, the influence of the proposed MRFP layer on running speed is small. The high running speed has huge advantages for real-world applications.

\begin{table}[t]
\centering
\resizebox{0.45\textwidth}{!}{%
\begin{tabular}{@{}llcc@{}}
\toprule
Method                         & GPU (TFLOPs)  & Image Size   & Speed (fps)   \\ \midrule
MGTS~\cite{chen2018person}     & K80 (8.7)     & 1500x900     & 0.8           \\
QEEPS~\cite{munjal2019query}   & P6000 (12.0)  & 1500x900     & 3.3           \\
NAE+~\cite{chen2020norm}       & V100 (14.1)   & 1500x900     & 10.2          \\
BINet~\cite{dong2020bi}        & 1080Ti (11.5) & 1500x900     & 12.5          \\
AlignPS~\cite{li2021sequential}& V100 (14.1)   & 1500x900     & 16.4          \\
DKD~\cite{zhang2021diverse}    & 1080Ti (11.5) & 1500x900     & 8.1           \\ \midrule
Baseline (STL)                 & V100 (14.1)   & 1500x900     & 30.5          \\
Baseline (STL)+MRFP            & V100 (14.1)   & 1500x900     & 28.7          \\ \midrule
Baseline (STL)+MRFP            & V100 (14.1)   & 960x640      & 40.6           \\ \bottomrule
\end{tabular}%
}
\caption{Running speed comparison.}
\label{tab:running_speed}
\end{table}
\section{Conclusion}
In this paper, we proposed the Subtask-dominated Transfer Learning (STL) method to address the long-tail identity distribution problem for the one-step person search. Compared to the widely-used ImageNet pretraining transfer learning method, the proposed STL method can significantly boost person search performance. 
Besides, we further proposed the Multi-level RoI Fusion Pooling (MRFP) layer to replace the single-level RoI pooling operation. The proposed MRFP layer can further improve the discriminative ability of the learned person features by keeping more details beneficial to the identification networks. Thorough experiments and ablation studies demonstrate the superiority and effectiveness of our proposed method.

{\small
\bibliographystyle{ieee_fullname}
\bibliography{egbib}
}

\end{document}